\pdfoutput=1

\documentclass[11pt]{article}

\usepackage[]{ACL2023}

\usepackage{times}
\usepackage{latexsym}
\usepackage{balance}
\usepackage{tabularx}

\usepackage[T1]{fontenc}

\usepackage[utf8]{inputenc}

\usepackage{microtype}

\usepackage{inconsolata}

\usepackage{graphicx}
\usepackage[caption=false,font=footnotesize]{subfig}
\usepackage{amsmath}
\usepackage{amsthm}
\usepackage{booktabs}
\usepackage{algorithm}
\usepackage{algorithmic}
\usepackage[caption=false,font=footnotesize]{subfig}
\usepackage{amsfonts}
\usepackage{multirow}
\usepackage{enumitem}
\usepackage{amssymb}
\usepackage{url}
\newcommand{\hide}[1]{}

\usepackage{newfloat}
\usepackage{tikz}
\usepackage{lipsum} 
\setlist[itemize]{leftmargin=*}

\newtheorem{hyp}{Hypothesis}

\usepackage{listings}

\title{Multidimensional Perceptron for Efficient and Explainable Long Text Classification}

\author{Yexiang Wang, Yating Zhang, Xiaozhong Liu, Changlong Sun  \\
        Alibaba Group \\
        \texttt{\{wangyexiang.wyx, ranran.zyt, xiaozhong.lxz\}@alibaba-inc.com,} \\
        \texttt{changlong.scl@taobao.com}}

\begin{document}
\maketitle
\begin{abstract}

Because of the inevitable cost and complexity of transformer and pre-trained models, efficiency concerns are raised for long text classification. Meanwhile, in the highly sensitive domains, e.g., healthcare and legal long-text mining, potential model distrust, yet underrated and underexplored, may hatch vital apprehension. Existing methods generally segment the long text, encode each piece with the pre-trained model, and use attention or RNNs to obtain long text representation for classification. In this work, we propose a simple but effective model, \textbf{S}egment-a\textbf{W}are mult\textbf{I}dimensional \textbf{PE}rceptron (SWIPE), to replace attention/RNNs in the above framework. Unlike prior efforts, SWIPE can effectively learn the label of the entire text with supervised training, while perceive the labels of the segments and estimate their contributions to the long-text labeling in an unsupervised manner. As a general classifier, SWIPE can endorse different encoders, and it outperforms SOTA models in terms of classification accuracy and model efficiency. It is noteworthy that SWIPE achieves superior interpretability to transparentize long text classification results.

\end{abstract}


\vspace{-6pt}
\section{Introduction}
\vspace{-6pt}


Text classification is a fundamental task in natural language processing (NLP). The performance of classification methods for short text has improved dramatically thanks to the advent of transformer-based pretrained language models (e.g., BERT). However, due to the shortage of transformer-based models that their self-attention operation scales quadratically with the text length, they are unable to process long sequences over 512 tokens. x
A simplest approach to solve this problem is to segment the input into smaller chunks and then add attention or RNNs to obtain an embedding for the entire input text for classification \cite{joshi2019bert}. In addition, Longformer \cite{beltagy2020longformer} was proposed to reduce the time complexity by simplifying the transformers to make the self-attention operation scale linearly with the sequence length.

\begin{figure}
    \centering
    \includegraphics[width=1\columnwidth]{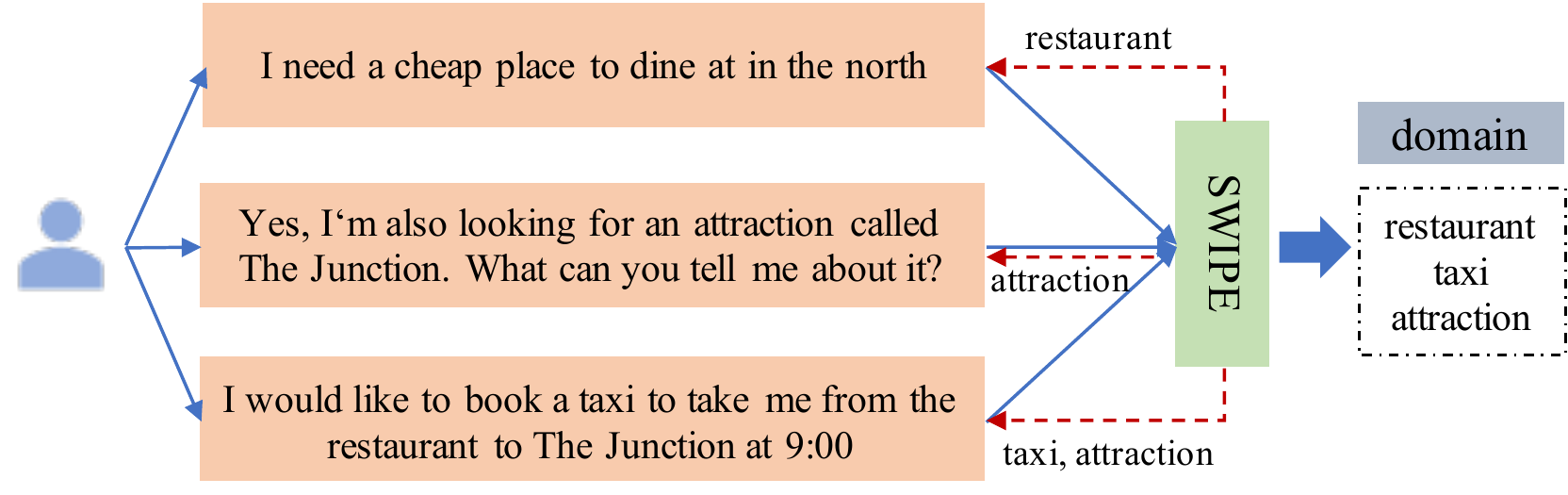}
    \vspace{-18pt}
    \caption{Example of explainable segment mining in multi-label classification task on dataset \textit{Multiwoz 2.2}.}
    \vspace{-18pt}
    \label{fig:example}
\end{figure}
Another challenge lies in the fact that when the text is long, it may not be able to transparentize/explain the model results well. Because humans can easily determine whether the result is correct for short texts, but difficult for long texts due to bigger noise and the requirement for longer and continuous information. Figure \ref{fig:example} shows an example of discovering and labeling the explainable turns/segments which are predictive to interpret the categorical labels of the entire dialogue, which helps to improve the trust-worthiness of the classification results in relatively noisy dialogue scenario. Meanwhile, in safety-critical applications, such as health diagnosis or criminal justice, explanations are useful for the employers of these systems such as doctors and judges, to better understand a system’s strengths and limitations and to appropriately trust and employ the system’s predictions \cite{camburu2020explaining}. Therefore, the interpretability is more important and more difficult for long text classification. In addition, regarding the granularity of interpretable text, high granularity is better than low granularity in long text scenarios. For example, in a long judicial article, a certain paragraph of the criminal process is essential to the judgment result. From the perspective of human cognitive process of decision making, only highlighting individual words/ngrams is difficult for humans to obtain effective information from these scattered words in a short time. Instead, mining interpretable sentences with continuous information can potentially provide more assistance for downstream tasks, such as passage retrieval task and active learning.

In view of the challenges above, we propose a simple but effective classifier named \textbf{S}egment-a\textbf{W}are mult\textbf{I}dimensional \textbf{PE}rceptron (\textbf{SWIPE}) to target both efficiency and explainable issues in long text scenarios. 
The design of \textbf{SWIPE} lies in two hypotheses: (1) \textbf{\textit{The category of long text can be determined by some key text fragments}}. This observation has been found in many research work \cite{ding2020cogltx, fiok2021text}, but the problem is how to find the key sentences in more effective and efficient way. Existing work utilize supervised way with human annotations resulting in high cost and low generality. In \textbf{SWIPE}, a shared perceptron with linear layers is set to make the classifier learn by itself the contribution of each segments during classification. 
(2) \textbf{\textit{It is unnecessary to maintain such long-distance attention in classification task}}. In QA task, the answers to the question may appear in different positions and are far away from each other, thus the transformer is applied to learn the attention of every word to all the others so as to keep the long-distance dependence. However in classification task, as the first hypothesis suggests, the attention can be conducted only within the segmented text. In this paper, we focus more on the way to effectively fuse the information of each segment for categorizing the entire text.

\begin{figure*}
    \centering
    \includegraphics[width=0.93\textwidth]{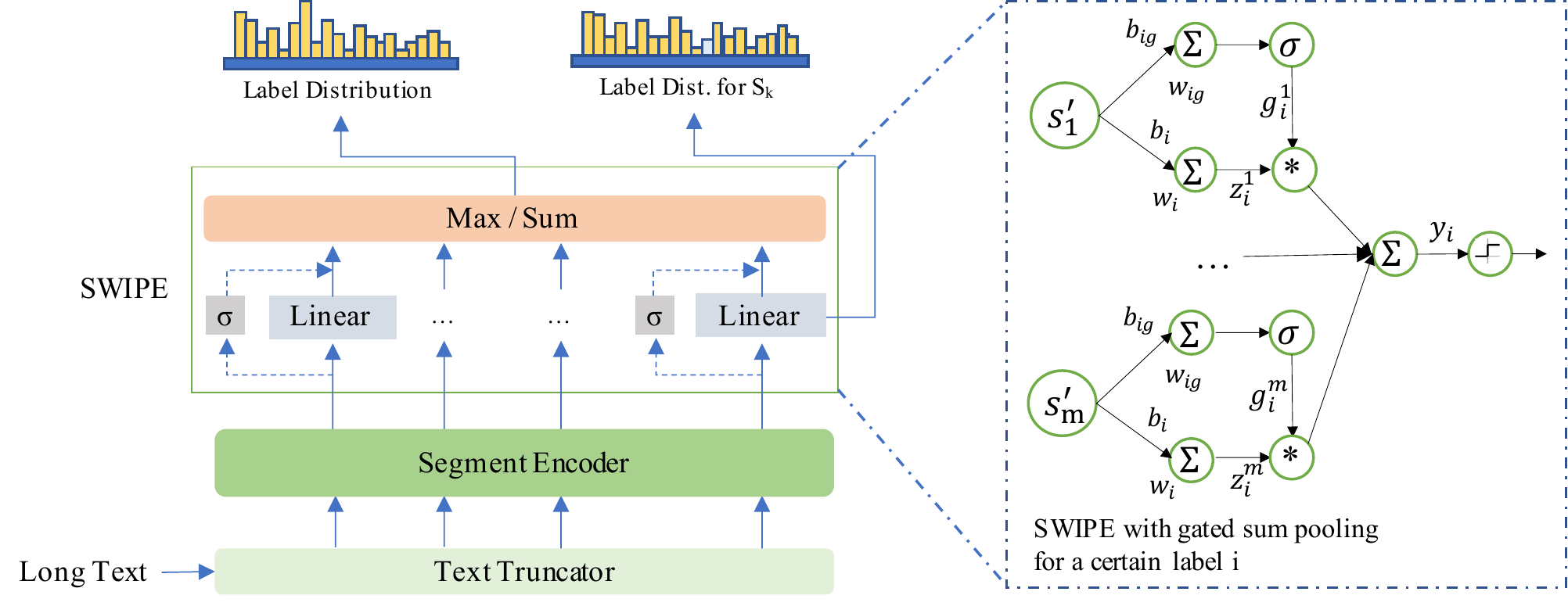}
    \vspace{-8pt}
    \caption{
    The model infrastructure of SWIPE for long text classification.}
    \vspace{-10pt}
    \label{fig:model}
\end{figure*}

Intuitively, \textbf{SWIPE} takes the segments of the long text as input, and assign each segment with a weight-sharing linear layer (see Figure \ref{fig:model}). The output of each linear layer is the label distribution of the corresponding segmented text and the final label of the entire text is determined by the label of its key sentence. The label distribution of the segmented text is then learned unsupervisedly through the training process of the classification of the entire text. Experiments over different type \& domain datasets demonstrate the superiority of \textbf{SWIPE} in terms of efficiency and explainability while maintaining competitive performance in accuracy, compared to the SOTA methods. Our contributions can be summarized as follows:
\begin{itemize}[noitemsep,nolistsep,topsep=0pt]
\setlength{\parskip}{0pt}
	\item We propose a novel classifier \textbf{SWIPE} which enables to categorize the text without length limitation and learn the key sentences by itself with good interpretability. 
	\item Experiments show the generality of \textbf{SWIPE} over different type \& domain datasets without discrimination to different encoders and meanwhile supporting both multi-class and multi-label classification tasks.
	\item The unsupervised interpretable process of \textbf{SWIPE} brings a major breakthrough in long text classification. To motivate other scholars, we make the code publicly available.
\end{itemize}

\section{Preliminaries}
\vspace{-4pt}

\subsection{Mathematical Foundation of Explainability of SWIPE}
\label{Sec:math_basis}
Consider a two-class ($1,0$) classification problem. 
Define document $d_A=\{s_a,s_b,s_c,s_d\}\in l$ and document $d_B=\{s_a,s_b,s_d,s_f\}\notin l$, where $s$ denotes the text segment in the document and document A is belonging to the target class $l$ and document B is not.
\vspace{-4pt}
\begin{hyp}
\label{hyp:hypothesis}
We assume that the label of a document is determined by the label of its most significant segment. 
\end{hyp}
\vspace{-4pt}
Based on this hypothesis, if document A belongs to category $l$ (expressed as $z(d_A)=wx_A+b>0$), there must be a segment $s$ in the document A that belongs to category $l$ (expressed as $z_s=wx_s+b>0$) where the function $z$ is the standard perceptron. We can define $z(d_A)=\text{max}(z_1,...,z_n)$ to locate the position of segment $s$. Then document A and B can be expressed as below:
\setlength{\abovedisplayskip}{6pt}
\setlength{\belowdisplayskip}{6pt}
\begin{align}
z(d_A)&=\text{max}(z_a,z_b,z_c,z_d)>0 \\
z(d_B)&=\text{max}(z_a,z_b,z_d,z_f)<0
\end{align}
Therein we can obtain,
\begin{align}
\label{Eq:max result}
z_c>0
\end{align}
namely, segment $s_c$ is the reason of document A belonging to the target class $l$.

Therefore, if we can design a perceptron which satisfies the above deduction, then the explainable segment can be easily obtained.

\vspace{-6pt}
\subsection{SWIPE v.s. Perceptron}
\vspace{-4pt}
Based on the hypothesis above, we design a Segment-aWare multIdimensional PErceptron (SWIPE), with max pooling. We compare the standard perceptron with our proposed SWIPE, and also show the generality of SWIPE.

\noindent Standard Perceptron:
\begin{align}
\text{P}(x)=\left\{\begin{matrix}
 1,& \text{if} \; \; wx+b>0\\ 
 0,& \text{else}
\end{matrix}\right.
\end{align}
SWIPE with max pooling:
\begin{align}
\text{SWIPE}(X)=\left\{\begin{matrix}
 1,& \text{if} \; \; \text{max}(wX+b)>0\\ 
 0,& \text{else}
\end{matrix}\right.
\end{align}
where $X=\{x_1,x_2,...,x_m\}$.

\vspace{2pt}
\noindent When $m=1$,
\vspace{-6pt}
\begin{align}
\text{SWIPE}(X)=\text{P}(x)
\end{align}

\vspace{-4pt}
\section{Methodology}
\vspace{-6pt}
The model infrastructure is shown in Figure~\ref{fig:model}, which consists of a text truncator, a segment encoder and a segment-aware multidimensional perceptron (SWIPE).
The text truncator divides the long text into multiple segments. 
The segment encoder is responsible for encoding the segment and obtaining the pooled vector representation of each segment. 
Then SWIPE can output labels for the entire text as well as the labels for each segment.
\vspace{-4pt}
\subsection{Text Truncator}
\vspace{-2pt}
\label{Sec:truncator}
Following the similar way of text truncation for long text in the previous work \cite{ding2020cogltx, fiok2021text}, we first split the long text into multiple segments. 
Due to the unique expressiveness of SWIPE, the length of truncated segment has impact on the final classification results. We thus analyze the influence of segment length in  Sec. \ref{sec:efficiency}.
Generally we adopt the following three truncation methods.

\textbf{Automatic truncation} captures fixed-length text with a sliding window which keeps a certain amount of text overlap between two segments. This is a general way utilized by previous work \cite{wang2019multi,ding2020cogltx} due to its simplicity but such coarse-grained truncation may result in the separation of an entire sentence\footnote{To make fair comparison with the SOTA methods, we apply the automatic truncation to all the tested methods on the datasets \textit{20 NewsGroups}, \textit{IMDB}, \textit{arXiv}. The details are described in Sec. \ref{sec:training-details}.}.

\textbf{Punctuation-based truncation} is based on the punctuation marks, such as ".", as signs to cut off the sentences. Since the complete sentence information is retained, it can be more accurately observe the identified the explainable segments (namely, important sentences) by SWIPE during the text classification process\footnote{To help the readers better understand the explainable results, we attached an exemplar file to show the identified explainable sentences of several examples in dataset \textit{IMDB}.}.

\textbf{Structure-based truncation} is performed according to the structural information of the text. For example, the turns in a dialogue can be regarded the segments when coping with conversational data.

Let us define $ r $ as the original long text. After truncation, the segments can be expressed as $\{s_1,s_2,...,s_m\} = \text{truncate}(r)$.

\vspace{-2pt}
\subsection{Segment Encoder}
\vspace{-4pt}
\label{sec:seg-encoder}

To encode each segment, we follow the common way to get the pooled vector representations from pretrained models, i.e. the vector representations of [CLS] token by feeding each segment into BERT:
\begin{align}
    s_1',s_2',...,s_m' &= \text{BERT}(s_1,s_2,...,s_m),  \\
    r' &= [s_1',s_2',...,s_m'] \in \mathbb{R}^{m \times h}, 
\end{align}
where $ s_k' \in R^h $ is the vector representation of the $k_{th}$ segment, $ r' $ denotes long text, and $h$ is the dimension of the hidden state of BERT.

Additionally, we can add additional transformer layers to allow segment vectors to interact, which is an optional module mainly used for comparison in our experiments:
\begin{align}
\label{Eq:transformer}
    s_1'',s_2'',...,s_m'' &= \text{T}(s_1',s_2',...,s_m'),
\end{align}


\vspace{-2pt}
\subsection{Segment-aware Multidimensional Perceptron (SWIPE)}
\vspace{-2pt}
\label{Sec:MDP}
SWIPE accepts multiple segment vectors and outputs a binary value for a certain label. The details are depicted in  Figure~\ref{fig:model}.

First, SWIPE processes each vector input as a normal perceptron:
\begin{align}
z_i^k &= w_i s_k' + b_i, 
\end{align}
where $ z_i^k \in R^h $ is the score of segment $ k $ on label $ i $ and $ w_i $ can be a randomly initialized weight in the neural network or a vector representation of the label $ i $.

At the same time, SWIPE calculates another score $ g_i^k $ for each segment to control the participation of this segment in the subsequent pooling:
\begin{align}
\label{Eq:gate}
g_i^k &= \sigma(w_i^g s_k' + b_i^g) \subseteq (0, 1), 
\end{align}
where $ w_i^g $ is a weight from random initialization.

Then, we can obtain the score $ y_i $ of the long text $ r $ on the label $i$ by one of the following pooling strategies:

\textit{Max Pooling}
\begin{align}
\label{Eq:max pooling}
y_i  = \text{max}(z_i^1, z_i^2, ..., z_i^m),
\end{align}

\textit{Gated Max Pooling}
\begin{align}
\label{Eq:gated max pooling}
y_i & = \text{max}(g_i^1 z_i^1, g_i^2 z_i^2, ..., g_i^m z_i^m), 
\end{align}

\textit{Sum Pooling}
\begin{align}
\label{Eq:sum pooling}
y_i & = \sum_{k=1}^m z_i^k, 
\end{align}

\textit{Gated Sum Pooling}
\begin{align}
\label{Eq:gated sum pooling}
y_i & = \sum_{k=1}^m g_i^k z_i^k.
\end{align}

Finally, the \textit{step function} can be used to determine whether the text $ r $ belongs to the label $i$:
\begin{align}
  b_i = \left\{\begin{matrix}
    1, & \text{if} \; \; y_i>0 \\
    0, & \text{else}
  \end{matrix}\right.
\end{align}

where text $r$  belongs to label $i$ when $b_i=1$.

We have introduced the mathematical foundation of \textit{Max Pooling} in \ref{Sec:math_basis}. 
Here we also provide the deduction for \textit{Sum Pooling} to show its consistency with Hypothesis \ref{hyp:hypothesis}.

Define a document A, $ Z_A = [z_a, z_b, z_c, z_d] $, a document B, $ Z_B = [z_a, z_c, z_d] $, and a label $ l $, where $ A \in l $ and $ B \notin l $. When adopting \textit{Sum Pooling}, it can be expressed as:
\setlength{\abovedisplayskip}{5pt}
\setlength{\belowdisplayskip}{5pt}
\begin{align}
z_a + z_b + z_c + z_d > 0, \\
z_a + z_c + z_d < 0,
\end{align}
then we can also get:
\begin{align}
\label{Eq:sum result}
z_b > 0,
\end{align}
which means the segment $ b $ is the reason of document A belonging to the target class $l$.

Therefore, no matter which pooling method is used, we can use $ z_i^k $ to determine whether segment $ k $ belongs to label $ i $ , 
\begin{align}
 \label{Eq:segment labeling}
  b_i^k = \left\{\begin{matrix}
    1, & \text{if} \; \; z_i^k>0 \\
    0, & \text{else}
  \end{matrix}\right.
\end{align}


where $ b_i^k = 1 $ means segment $ k $  belongs to label $ i $.
In addition, $ z_i^k $ or $ g_i^k z_i^k $ can also be used to rank segments for label-related relevance. 

Note that although \textit{Max} and \textit{Sum} poolings are supported by the same hypothesis, but for \textit{Sum Pooling}, the key segment only influences the final result rather than determining the final result as \textit{Max Pooling} does, due to the difference between the two operations. Details of their difference are described in Sec. \ref{sec:ablation}. 

\vspace{-2pt}
\section{Experimental Settings}
\vspace{-4pt}
\begin{table}[!t]
\centering
\scriptsize
\begin{tabularx}{\columnwidth}{lccccc}
\hline
                      & \textbf{\# Train}  & \textbf{\# Dev } & \textbf{\# Test} & \textbf{Len$_{avg}$} & \textbf{Len$_{max}$} \\ \hline
\textit{20 NewsGroups}       & 10,741      & 573              & 7,532            & 258                  & 11,154               \\
\textit{IMDB full}           & 22,500      & 2,500            & 25,000           & 231                  & 2,470                \\
\textit{IMDB Long}           & 1,815       & 203              & 1,885            & 689                  & 2,470                \\
\textit{arXiv}      & 15,988      & 3,192            & 12,781           & 10,032               & 552,706              \\
\textit{MultiWOZ 2.2}        & 8,324       & 1,000            & 1,000            & 80                   & 340                  \\ \hline
\end{tabularx}
\vspace{-4pt}
\caption{Statistics of datasets. Len$_{avg}$ denotes the average number of words per document, and Len$_{max}$ is the maximum number of words. }
\vspace{-15pt}
     \label{tab:dataset_statistics}
\end{table}



    











\subsection{Dataset}
\label{sec:dataset}
To validate the performance of \textbf{SWIPE} on long text classification in terms of accuracy, efficiency and explainability, we adopt several public datasets from various domains as benchmark.

\textit{20 NewsGroups} \cite{lang1995newsweeder} is a multi-class text classification dataset.
We clean the data by removing formatting characters to preserve only text messages, which is consistent with the setting in previous work \citep{ding2020cogltx}.
\textit{ArXiv Data Set} \cite{he2019long} is a multi-label classification dataset collected from arXiv web site. 
\textit{IMDB} \cite{maas2011learning} is a binary sentiment classification review dataset collected from IMDB.
Due to the varying length of different reviews, we further prepare an addition dataset (denoted as \textit{IMDB Long}) by selecting the reviews over $500$ words.
\textit{MultiWOZ 2.2} \cite{zang2020multiwoz} is a multi-label classification dialogue dataset, where the labels of the dialogue is the set of labels assigned to its key utterances. In the experiment, we regard the dialogue as document and the turn as segment.
\vspace{-4pt}
\subsection{Evaluation Metrics} 
\label{Sec:evaluation_metrics}
\vspace{-2pt}
We utilize accuracy as the evaluation metric for binary and multi-class classification tasks (datasets \textit{IMDB} and \textit{20 NewsGroups}). As for multi-label classification (\textit{arXiv} and \textit{MultiWOZ 2.2}), Micro F$_1$ and Macro F$_1$ scores are employed to evaluate the classification performance.

To quantitatively justify the explainability of the classification model, we follow the setting in \cite{rajagopal2021selfexplain} by adopting the criterion \textbf{Sufficiency} which aims to evaluate whether model explanations (namely the extracted explainable text segments) alone are highly indicative of the predicted label \cite{jacovi2018understanding,yu2019rethinking}. To evaluate sufficiency of explanations, we employ fasttext \cite{joulin2017bag} as classifier trained to perform a task using only the extracted explanations without the rest of the input. The explanation that achieves high accuracy using this classifier is indicative of its ability to recover the original model prediction.

In addition, in dataset \textit{MultiWOZ 2.2}, there exists annotated key/explainable segment for each dialogue, thus the result (Micro F1 and Macro F1) achieved by supervised method of explainable segment labeling can be regarded as an upper bound of \textbf{SWIPE} thanks to its unsupervised manner in locating key sentences.


\subsection{Training Details}
\vspace{-4pt}
\label{sec:training-details}
In the experiments, the pretrained language models are running on their base version. All the models are trained in $10$ epochs to minimize cross-entropy loss with the Adam optimizer \cite{kingma2014adam} in which the learning rate linearly decreases from $5e^{-5}$. Under this setting, for each task, we report the average result on test set over five random initializations without model ensemble.

\begin{table}[t]
    \centering
    \small
    \begin{tabular}{clc}
    \hline
                         & \textbf{Model}             & \textbf{Accuracy}  \\
    \hline
    \multirow{3}{*}{(1)} & fastText                   & 79.4  \\
                         & MS-CNN                     & 86.1  \\
                         & TextGCN                    & 86.3  \\
    \hline
    \multirow{3}{*}{(2)} & BERT                       & 87.2  \\
                         & RoBERTa                    & 86.5  \\
                         & Albert                     & 84.6  \\
    \hline
    \multirow{2}{*}{(3)} & BERT + attention           & 87.0 \\
                         & BERT + LSTM                & 84.7 \\
    \hline
    \multirow{2}{*}{(4)}                  & Longformer                 & 86.6 \\
                      & CogLTX                     & 87.0 \\
    \hline
                         & BERT + SWIPE$_{sum}$ + gate  & 87.3 \\
                         & BERT + SWIPE$_{sum}$ + gate + 2t      & \textbf{87.4} \\
    \hline
    \end{tabular}
    \vspace{-4pt}
    \caption{Classification results on \textit{20 NewsGroups} over all the baselines and the best performing SWIPE variants.}
    \vspace{-12pt}
     \label{tab:main-20news}
\end{table}

\begin{table}[]
\centering
\small
\begin{tabular}{llc}
\hline
\textbf{Segment Len.}      & \textbf{Model}       & \textbf{Accuracy} \\ \hline
\textbf{-}                   & fasttext             & 82.44             \\ \hline
\multirow{4}{*}{64}          & random               & 64.35             \\
                             & LIME                 & 64.88   \\
                             & attention            & 77.24             \\
                             & SWIPE                & \textbf{81.75}    \\ \hline
\multirow{4}{*}{128}         & random               & 68.91             \\
                             & LIME                 & 68.97             \\
                             & attention            & 77.19             \\
                             & SWIPE                & \textbf{81.06}    \\ \hline
\multirow{4}{*}{256}         & random               & 72.31             \\
                             & LIME                 & 73.21             \\
                             & attention            & 77.93             \\
                             & SWIPE                & \textbf{82.02}    \\ \hline
\multirow{4}{*}{512}         & random               & 76.39             \\
                             & LIME                 & 78.09             \\
                             & attention            & 79.42             \\
                             & SWIPE                & \textbf{81.01}    \\ \hline
\end{tabular}
\caption{Sufficiency Test: Model predictive performances (prediction accuracy) on \textit{IMDB Long} test set. The model \textbf{Random}, \textbf{LIME}, \textbf{attention} and \textbf{SWIPE} refer to the models trained only with the selected segments by employing different ways of explanation technique.}
    \vspace{-14pt}
\label{tab:sufficiency}
\end{table}

\subsection{Tested Methods}
\vspace{-2pt}
To extensively validate the effectiveness of \textbf{SWIPE} in classification performance, the following four groups of baselines that address the problem of long text classification from different perspectives are employed for comparison.

\noindent $(1)$ \textit{Classic text classification models}

\textbf{fastText} \cite{joulin2017bag}, \textbf{MS-CNN} \cite{pappagari2018joint} and \textbf{TextGCN} \cite{yao2019graph} are the three classic text classification models to be compared in the experiments.

\noindent $(2)$ \textit{Standard pretrained language models}

\textbf{BERT} \cite{DBLP:conf/naacl/DevlinCLT19-bertraw}, \textbf{RoBERTa} \cite{liu2019roberta} and \textbf{Albert} \cite{lan2019albert} are the widely adopted pretrained language models (PLM). 

\noindent $(3)$ \textit{Truncated pretrained language models}

\textbf{BERT+attention} \cite{sun2020attention} and \textbf{BERT+LSTM} \cite{pappagari2019hierarchical} segment the input text into pieces, encode each piece of text with the pretrained model, and then use attention or LSTM to obtain an embedding for the entire input text for classification.

\noindent $(4)$ \textit{SOTA models for long text classification}

\textbf{Longformer} \cite{beltagy2020longformer} simplifies the self-attention with an attention mechanism that scales linearly with sequence length.
\textbf{CogLTX} \cite{ding2020cogltx} optimizes the way of sentence selection by training a judge model and concatenate them for reasoning.

To further justify the explainability of \textbf{SWIPE}, the classical explanation technique \textbf{LIME} \cite{ribeiro2016should} with random forest \cite{breiman2001random} is employed for comparison.

\begin{table*}[!t]
\centering
\small
\begin{tabular}{lcccc}
\hline
                   & \textit{IMDB Full}   & \textit{IMDB Long}   & \multicolumn{2}{c}{\textit{arXiv }}          \\ \cline{2-5} 
\textbf{Model}     & \textbf{Accuracy}    & \textbf{Accuracy}    & \textbf{Micro F1}    & \textbf{Macro F1}    \\ \hline
Longformer$_{4096}$         & 95.04                & 92.10                & 81.33                & 80.75                \\ \hline
BERT$_{512}$           & 93.91                & 88.12                & 79.94                & 79.37                \\
BERT + SWIPE$_{max}$    & 94.12                & 90.19                & \textbf{82.76}       & \textbf{82.18}       \\
RoBERTa$_{512}$      & 94.94                & 91.14                & 79.86                & 79.28                \\
RoBERTa + SWIPE$_{max}$ & \textbf{95.16}       & \textbf{92.52}       & 82.47                & 81.95                \\
Albert$_{512}$       & 93.02                & 88.12                & 79.33                & 78.83                \\
Albert + SWIPE$_{max}$  & 93.68                & 89.92                & 81.76                & 81.76                \\ \hline
                   & \multicolumn{1}{l}{} & \multicolumn{1}{l}{} & \multicolumn{1}{l}{} & \multicolumn{1}{l}{}
\end{tabular}
\vspace{-12pt}
\caption{Classification results on extreme long text datasets.}
\vspace{-10pt}
\label{tab:extreme-long_imdb+arxiv}
\end{table*}

In this work, we mainly test the variants of SWIPE with the settings of max pooling (see Eq.\ref{Eq:max pooling}) and sum pooling (see Eq.\ref{Eq:sum pooling}) together with different PLM based encoders. In addition, the components - the gate function (see Eq.\ref{Eq:gate}) and additional transformer layers (see Eq.\ref{Eq:transformer}) are added to different pooling strategy for testing. The notation of the variant is further expressed as ``PLM + SWIPE pooling strategy + component'', for instance \textbf{BERT+SWIPE$_{max}$+gate} denotes SWIPE with gated max pooling conducted on Bert based encoder.

\section{Result Analysis}

\begin{table}[!t]
\small
\centering
\begin{tabular}{lcc}
\hline
\textbf{Model}           & \textbf{Micro F1}            & \multicolumn{1}{l}{\textbf{Macro F1}} \\ \hline
BERT   (supervised)      & 85.55                        & 85.32                                 \\ 
BERT +  attention        & 62.38                        & 62.97                                 \\ \hline
BERT + SWIPE$_{max}$       & 73.99                        & 74.75                                 \\
\quad \textit{+ gate}    & 73.50                        & 74.34                                 \\
\quad \textit{+ 2t}      & 77.41                        & 77.53                                 \\
\quad \textit{+ gate + 2t}  & 77.59                     & 77.72                                 \\ \hline
BERT + SWIPE$_{sum}$       & 80.01                        & 79.21                                 \\
\quad \textit{+ gate}    & 81.13                        & 81.00                                 \\
\quad \textit{+ 2t}      & 67.89                        & 66.56                                 \\
\quad \textit{+ gate + 2t}  & \textbf{81.33}            & \textbf{82.16}                        \\ \hline
\end{tabular}
\vspace{-4pt}
\caption{Results of explainable segment labeling on the dataset \textit{MultiWOZ 2.2}.}
\vspace{-8pt}
\label{tab:explainability-multiwoz}
\end{table}

\vspace{-4pt}
\subsection{Overall Performance}
\vspace{-2pt}
$\bullet$ \textbf{Comparison with baselines.}
Table \ref{tab:main-20news} shows the performance of all the tested methods including the best variants of our proposed method\footnote{All the combinations of the proposed components have been tested and only the ones with best results are shown in Table \ref{tab:main-20news}.} on the dataset \textit{20 NewsGroups}. It is not surprising to observe the limited performance of the three classic text classification models compared with the other PLM based methods. Since the average length of the documents in \textit{20 NewsGroups} dataset is only $258$, therefore, the results do not differ much among various PLM based models. Under this circumstance, the variants of \textbf{SWIPE}\footnote{Here $2t$ in \textbf{BERT+SWIPE$_{sum}+2t$} denotes the variant by adding two transformation layers.} are still able to surpass all the baselines. In addition, in terms of the complexity of model implementation, for instance compared with the method \textbf{CogLTX}, \textbf{SWIPE} is also advantageous due to its simplicity and efficiency.
Due to the length limit, we only show the performance of all the baselines on the classic \textit{20 NewsGroups} dataset and the best baselines are further to be compared on the other datasets. 

\noindent$\bullet$ \textbf{Performance over extreme long text.}
When it comes to the scenario of extreme long text (the length of the document is over $500$ words) (see the results shown in Table \ref{tab:extreme-long_imdb+arxiv}), standard PLMs (\textbf{BERT}, \textbf{RoBERTa}, \textbf{Albert}) and \textbf{Longformer} start to show limited performance and the proposed \textbf{SWIPE} outperforms the selected baselines at a bigger margin. Especially in the case of \textit{arXiv} dataset with the average length over $10,000$ words, there exists significant superiority of the best variant \textbf{BERT+SWIPE$_{max}$} compared to the best baseline \textbf{Longformer}.

\vspace{-2pt}
\subsection{Explainability of SWIPE}
We evaluate the explanations on \textit{IMDB Long} dataset by adopting Sufficiency criterion as illustrated in Sec. \ref{Sec:evaluation_metrics}. We compare the predictive accuracy of the explanations from \textbf{SWIPE} in comparison to three baseline explanation methods. From Table \ref{tab:sufficiency}, we observe that explanations from \textbf{SWIPE} show higher predictive performance compared to all the baseline methods over different settings of segment length. We also notice the better performance of \textbf{Attention} (self-explanatory model) compared to 
\textbf{LIME}\footnote{We use \textbf{random forest+LIME} to select 6 keywords (from 1 to 10, 6 performed best) for each sample, then segment the original text in the same way as \textbf{SWIPE}, and select the segment containing the most keywords as the interpretable segment. } which belongs to post-hoc explanations. Meanwhile, SWIPE explanations achieves competitive performance with full-text setting (an explanation that uses all of the input sample) performance, indicating higher sufficiency of explanations. 

Moreover, the explainability of \textbf{SWIPE} can be quantitatively evaluated by the task of segment labeling using the dataset \textit{MultiWOZ 2.2}. Table \ref{tab:explainability-multiwoz} shows the results of the supervised training using \textbf{BERT} which is regarded as the upper bound and the ones trained unsupervisedly by the variants of \textbf{SWIPE}. 
The best performing variant \textbf{BERT+SWIPE$_{sum}$+gate+2t} can be comparable to the supervised upper bound. 
In addition, we can observe the poor performance of the baseline method 
\textbf{BERT+attention}\footnote{The results of the experiment are obtained after we set $1$ as the best threshold. Usually the attention score varies from $-10$ to $10$.} demonstrates the claim of ``\textit{Attention cannot be an Explanation}'' \cite{akula2022attention}.

\vspace{-4pt}
\subsection{Pluggability of SWIPE}
\vspace{-4pt}
To validate the pluggability of 
\textbf{SWIPE}, we conduct experiments by plugging the component of SWIPE with max pooling (\textbf{SWIPE$_{max}$}) on the three standard PLMs (\textbf{BERT}, \textbf{RoBERta} and \textbf{Albert}). According to the results shown in Table \ref{tab:extreme-long_imdb+arxiv} (comparing the results at rows 3, 5, and 7 with the ones at rows 2, 4, and 6), the addition of \textbf{SWIPE$_{max}$} enables to enhance the performance of the vanilla PLMs to different extent. Specifically it brings about an average of $0.4\%$ improvement on regular length dataset (\textit{IMDB Full}), and an average growth of $2\%$ and $3.5\%$ on the extreme long text datasets \textit{IMDB Long} and \textit{arXiv} respectively. On \textit{IMDB}, \textbf{RoBERTa+SWIPE$_{max}$} performs best while \textbf{BERT+SWIPE$_{max}$} outperforms the other variants on \textit{arXiv}. Therefore, we can generalize the effectiveness of \textbf{SWIPE} without discrimination to different encoders. Moreover, it can be seen that the longer the text, the more improvement the \textbf{SWIPE} can bring about.

\begin{table}[!t]
\centering
\begin{tabular}{lc}
\hline
\textbf{Model}  & \textbf{Accuracy} \\ \hline
BERT + SWIPE$_{max}$ & 86.51             \\
\quad \textit{+ gate}           & 86.30             \\
\quad \textit{+ 2t}           & 86.76             \\
\quad \textit{+ gate + 2t}     & 87.04             \\ \hline
BERT + SWIPE$_{sum}$ & 86.68             \\
\quad \textit{+ gate}          & 87.27             \\
\quad \textit{+ 2t}            & 86.96             \\
\quad \textit{+ gate + 2t}     & \textbf{87.41}    \\ \hline
\end{tabular}
\vspace{-4pt}
\caption{Ablation Study}
\vspace{-18pt}
\label{tab:ablation}
\end{table}

\vspace{-2pt}
\subsection{Ablation Study}
\label{sec:ablation}
In this section, we discuss the effect of different combinations of the proposed components in \textbf{SWIPE}. Table \ref{tab:ablation} shows the results of SWIPE variants conducted on the dataset \textit{20 NewsGroups}. In general, we can see the increase in the performance by adding the two components - gate and transformation layers to the basic \textbf{SWIPE$_{max}$} and \textbf{SWIPE$_{sum}$}. 
To be specific, the gate function is more suitable for sum pooling since the weighted summation enables to learn which segment is more important. In addition, since the truncation of text have lost the interactions between the segments, the transformation layers thus make up for this loss. We have conducted experiments with different number of transformation layers and it appears that two layers achieve the best results. However adding transformation layers will result in the increase of memory and time complexity, therefore whether it should be added to the model should be determined for practical use.   

Moreover, to compare the results of the variants based on \textbf{SWIPE$_{max}$} and the ones based on \textbf{SWIPE$_{sum}$}, the latter ones perform slightly better than the former ones. Note that Hypothesis \ref{hyp:hypothesis} works for both of the two pooling strategies which can be supported by Eq. \ref{Eq:max result} and Eq. \ref{Eq:sum result}. But for \textbf{SWIPE$_{sum}$}, the key segment only influences the final result rather than determining the final result as \textbf{SWIPE$_{max}$} does, due to the difference between the two operations, summation and taking the maximum. 
\begin{figure}
    \centering
    \includegraphics[width=0.95\columnwidth]{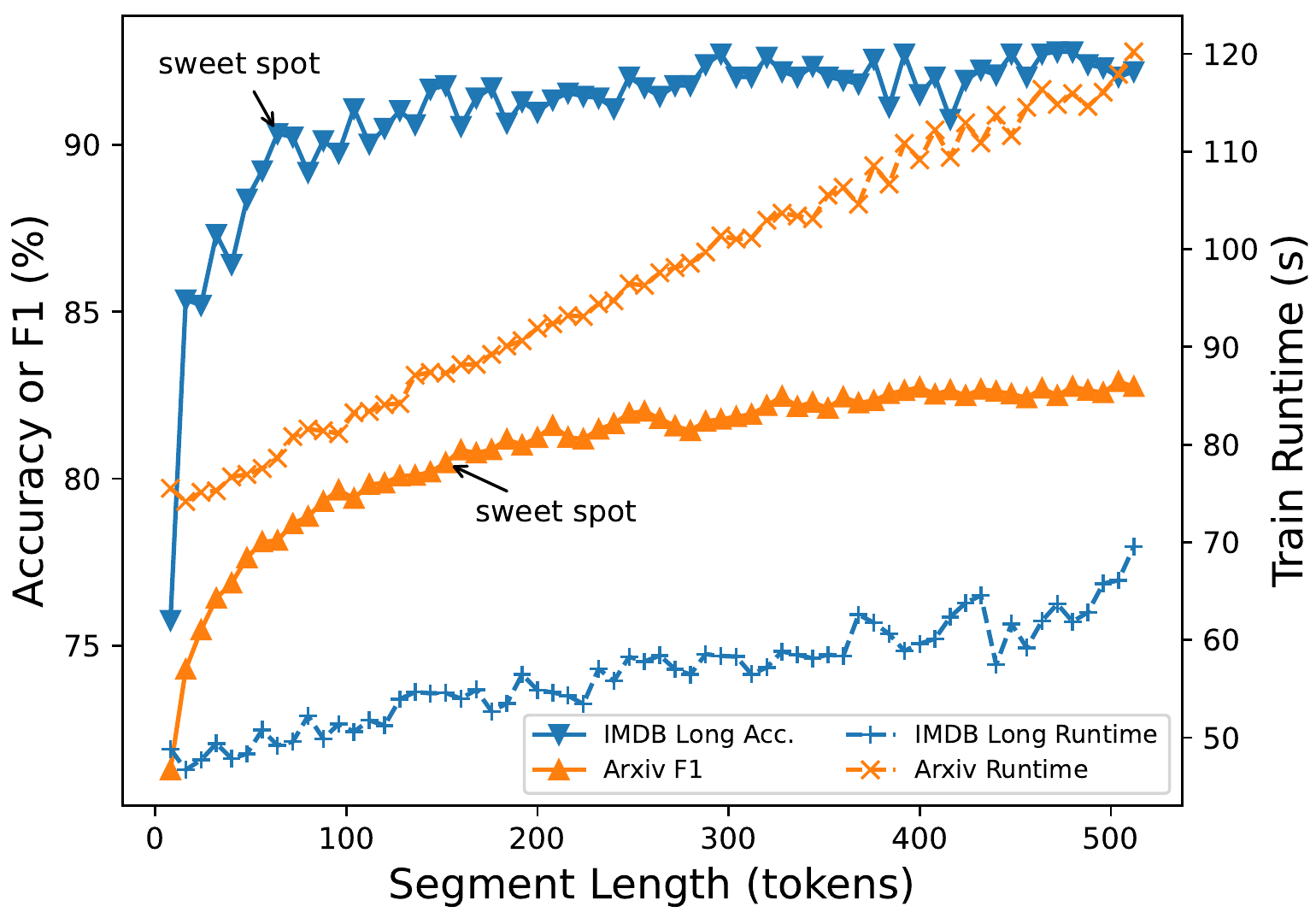}
     \vspace{-4pt}
    \caption{
    Effectiveness and efficiency analysis of SWIPE. 
    Since the scale of \textit{arXiv} is much larger than \textit{IMDB Long}, we reduce the train runtime of  \textit{arXiv} by a factor of 1000 in the figure.
    }\label{fig:efficiency}
    \vspace{-12pt}
\end{figure}

\vspace{-2pt}
\subsection{Efficiency of SWIPE}
\vspace{-4pt}
\label{sec:efficiency}
The time complexity of basic \textbf{SWIPE} is $O(dnm)$ where $d$ represents the dimension of embedding, $n$ denotes the length of the document and $m$ is the length of each segment. When $m$ is fixed, the time complexity of \textbf{SWIPE} is linearly increasing with the length of document. In this section, we also investigate the impact of the segment length ($m$) on the corpus with different length in terms of the accuracy and training time. Figure \ref{fig:efficiency} shows the performance and efficiency of model \textbf{SWIPE$_{max}$} on the datasets \textit{IMDB full} and \textit{arXiv}. 

As for the training time, it presents a linear relationship with the increase of segment length when the length of document $n$ is fixed. On the other hand, the accuracy increases with the growth of the segment length. However, we can observe a ``sweet spot'' where the accuracy rises slowly after reaching this spot. Experimentally, we select this spot by taking the $97\%$ of the best accuracy.
Compared with the two datasets, the ``sweet spot'' for \textit{arXiv} (152 tokens) is larger than that for \textit{IMDB} (64 tokens) due to the longer text in \textit{arXiv}. For practical use, considering the training cost, selecting an appropriate segment length is essential for the scenario of high efficiency.

\begin{figure}
    \centering
    \includegraphics[width=0.95\columnwidth]{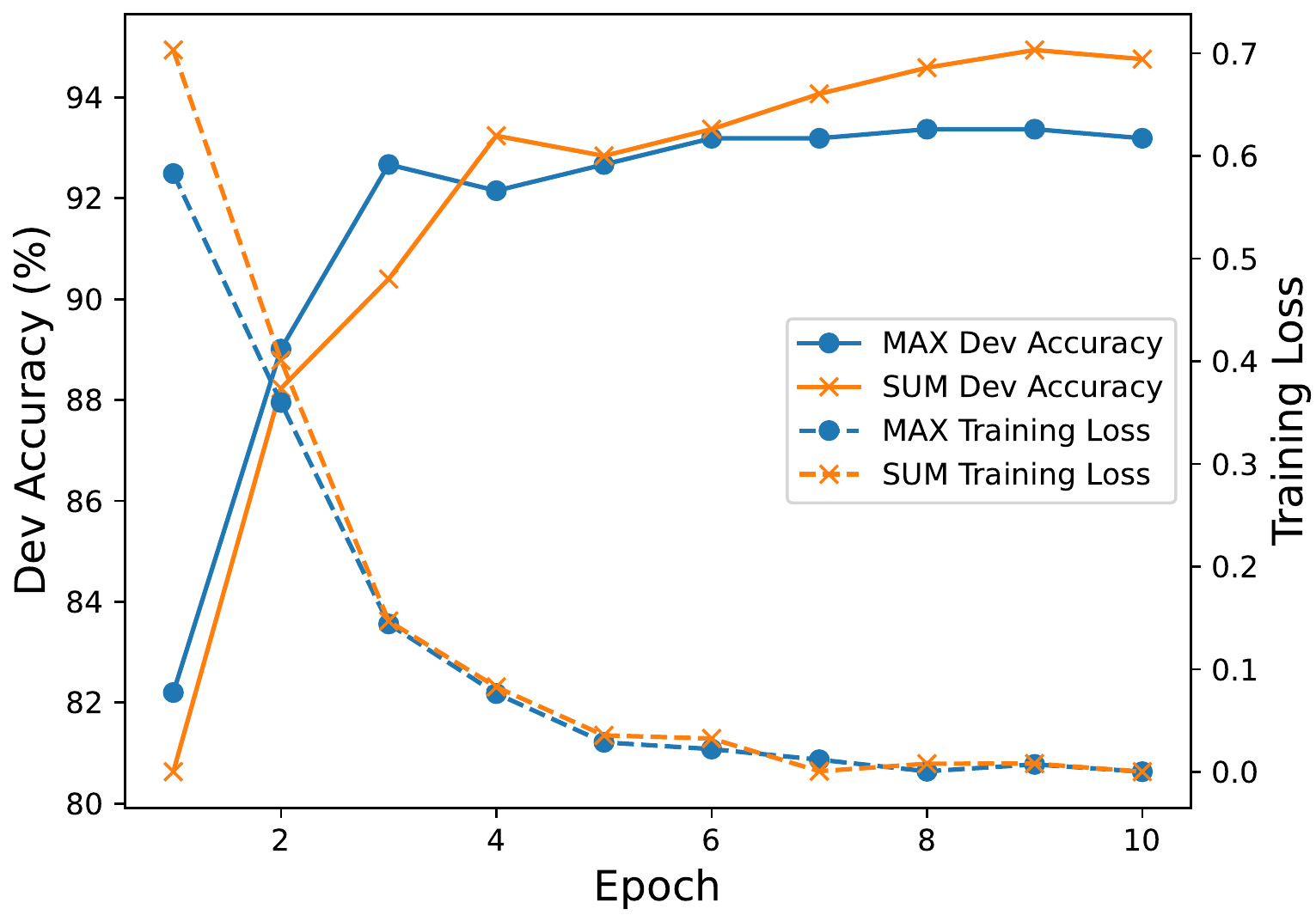}
     \vspace{-4pt}
    \caption{
    Convergence analysis of SWIPE on \textit{20 NewsGroups} dataset.
    }\label{fig:convergence}
    \vspace{-12pt}
\end{figure}

\subsection{Convergence Analysis}
\vspace{-2pt}
To better understand the learning process of the variants \textbf{SWIPE$_{max}$} and \textbf{SWIPE$_{sum}$}, we conduct experiments to monitor the impact of different pooling strategies on the convergence of long text classification task. As depicted in Figure \ref{fig:convergence}, in the first few epochs, \textbf{SWIPE$_{max}$} achieves better accuracy than \textbf{SWIPE$_{sum}$}. Correspondingly, \textbf{SWIPE$_{max}$} converges faster at the beginning. However \textbf{SWIPE$_{sum}$} can obtain higher accuracy after continuous training.

\section{Related Work}
\vspace{-4pt}
\subsection{Long Text Classification}
\vspace{-2pt}
Deep learning methods with pretrained language models have achieved good results in short text classifications \cite{lee2016sequential, wang2017combining}. When it comes to the long text scenarios, the transformer-based pretrained models (e.g., BERT) are incapable of processing long texts due to its quadratically increasing memory and time consumption. The most natural ways to address this problem are generalized as: truncating the text into segments \cite{joshi2019bert, xie2020unsupervised,wang2019multi} or simplifying transformers \cite{beltagy2020longformer}. 
In this work, we solve this problem from a novel perspective by introducing a simple but effective classifier \textbf{SWIPE}. Extensive experiments demonstrate its efficiency and generality over different type and domain datasets.


\vspace{-4pt}
\subsection{Explainability of Text Classification}
\vspace{-2pt}

Post-hoc explanatory methods are stand-alone methods that aim to explain already trained and fixed target models \cite{camburu2020explaining}.  For example, LIME \cite{ribeiro2016should} explains a prediction of a target model by learning an interpretable model. 
Self-explanatory models incorporate an explanation generation module into their architecture such that they provide explanations for their own predictions 
Existing approaches. Existing work \cite{arras2017relevant, samek2017explainable} often rely on heat maps based on attention to visualize the most relevant terms to the predicted results. 
Another recent line of work explores collecting rationales \cite{lei2016rationalizing,arous2021marta,liu2019towards} through expert annotations to learn the relevance of the input text to the predicted results. However such approach is difficult to generalize to other domains due to high manual cost. 
\textbf{SWIPE} is self-explanatory model and the interpretable sentences are learned in an unsupervised manner during the classification process without extra effort thanks to its mathematical property.

\vspace{-4pt}
\section{Conclusion}
\vspace{-4pt}
In this paper, we propose a simple but effective classifier called segment-aware multidimensional perceptron  which enables to categorize the text without length limitation and learn the key segments by itself with good interpretability. Thanks to the mathematical property of SWIPE, it can learn the labels of each segments and their contributions to the entire text label in an unsupervised manner. As a general classifier without discrimination to different encoders, experiments demonstrate the superiority of SWIPE in terms of classification accuracy and model efficiency. The unsupervised interpretable process brings a major breakthrough in long text classification, allowing for better optimization of model performance in the future.
\section*{Limitations}
In this section, we discuss the limitations of our work as follows:
\begin{itemize}
    \item The length of segments has impact on the accuracy of text classification of SWIPE. When the truncation of segments is unreasonable, such as very short, the performance of SWIPE will drop significantly, as shown in Figure \ref{fig:efficiency}. 
    \item In Sec. \ref{Sec:truncator} and Sec. \ref{sec:seg-encoder}, we have suggested the way of encoding long text by truncating the text into segments and encoding each segment using PLM (e.g., BERT). However we also tried to encode the entire piece of text with BERT and then utilize the vector of each token as the input to SWIPE for classification. Although the accuracy of text classification for the entire text stayed the same, the explainability of SWIPE, however, was lost. It is due to the strong fitting capability of BERT. When combined with SWIPE, it makes the embeddings of all the tokens similar to each other, thus losing their own characteristics.
\end{itemize}


\bibliography{Main}
\bibliographystyle{acl_natbib}


\end{document}